%% file: main.tex
\documentclass[sigconf]{acmart}

\usepackage{algorithm}
\usepackage[noend]{algpseudocode}

\usepackage{newfloat}
\usepackage{listings}
\usepackage{footmisc}
\usepackage{microtype}
\usepackage{latexsym}
\usepackage{xspace}
\usepackage{booktabs}
\usepackage{multirow}
\usepackage{makecell}
\usepackage{color}
\usepackage{xcolor}
\usepackage{subcaption}
\usepackage{paralist}
\usepackage{cleveref}

\usepackage{enumitem}

\usepackage{soul}
\colorlet{soulred}{red!20}
\colorlet{soulblue}{cyan!30}
\DeclareRobustCommand{\hr}[1]{{\sethlcolor{soulred}\hl{ #1 }}}
\DeclareRobustCommand{\hb}[1]{{\sethlcolor{soulblue}\hl{ #1 }}}

\newcommand{\xhdr}[1]{\vspace{1.5ex}\noindent\textbf{#1}}

\usepackage[normalem]{ulem}
\useunder{\uline}{\ul}{}
\usepackage{pifont}
\newcommand{\cmark}{\ding{51}}%
\newcommand{\xmark}{\ding{55}}%
\newcommand{\our}{\mbox{\textsc{UCEpic}}\xspace}

\AtBeginDocument{%
  \providecommand\BibTeX{{%
    \normalfont B\kern-0.5em{\scshape i\kern-0.25em b}\kern-0.8em\TeX}}}

\copyrightyear{2023}
\acmYear{2023}
\setcopyright{rightsretained}
\acmConference[KDD '23]{Proceedings of the 29th ACM SIGKDD Conference on Knowledge Discovery and Data Mining}{August 6--10, 2023}{Long Beach, CA, USA}
\acmBooktitle{Proceedings of the 29th ACM SIGKDD Conference on Knowledge Discovery and Data Mining (KDD '23), August 6--10, 2023, Long Beach, CA, USA}
\acmDOI{10.1145/3580305.3599535}
\acmISBN{979-8-4007-0103-0/23/08}

\begin{document}

\title{UCEpic: Unifying Aspect Planning and Lexical Constraints \\ for Generating Explanations in Recommendation}



\author{Jiacheng Li}
\authornote{Both authors contributed equally to this research.}
\affiliation{%
  \institution{University of California, San Diego}
  \state{California}
  \country{USA}
}
\email{j9li@eng.ucsd.edu}

\author{Zhankui He}
\authornotemark[1]
\affiliation{%
  \institution{University of California, San Diego}
  \state{California}
  \country{USA}
}
\email{zhh004@eng.ucsd.edu}

\author{Jingbo Shang}
\affiliation{%
  \institution{University of California, San Diego}
  \state{California}
  \country{USA}
}
\email{jshang@ucsd.edu}

\author{Julian McAuley}
\affiliation{%
  \institution{University of California, San Diego}
  \state{California}
  \country{USA}
}
\email{jmcauley@ucsd.edu}

\renewcommand{\shortauthors}{Jiacheng and Zhankui, et al.}

\input{0_abstract}





\maketitle
\input{1_intro}
\input{2_related}
\input{3_method}
\input{4_experiments}
\input{5_conclusion}
\appendix
\input{6_appendix}

\bibliographystyle{ACM-Reference-Format}
\balance
\bibliography{main}

\end{document}

%% file: 0_abstract.tex
\begin{abstract}
Personalized natural language generation for explainable recommendations plays a key role in justifying why a recommendation might match a user's interests. Existing models usually control the generation process by aspect planning. While promising, these aspect-planning methods struggle to generate specific information correctly, which prevents generated explanations from being convincing. In this paper, we claim that introducing lexical constraints can alleviate the above issues. We propose a model, \our, that generates high-quality personalized explanations for recommendation results by unifying aspect planning and lexical constraints in an insertion-based generation manner. 

Methodologically, to ensure text generation quality and robustness to various lexical constraints, we pre-train a non-personalized text generator via our proposed robust insertion process. Then, to obtain personalized explanations under this framework of insertion-based generation, we design a method of incorporating aspect planning and personalized references into the insertion process. 
Hence, \our unifies aspect planning and lexical constraints into one framework and generates explanations for recommendations under different settings.
Compared to previous recommendation explanation generators controlled by only aspects, \our incorporates specific information from keyphrases and then largely improves the diversity and informativeness of generated explanations for recommendations on datasets such as RateBeer and Yelp.

\end{abstract}

%% file: 1_intro.tex
\begin{table}[thbp]
\centering
\small
\caption{Comparison of previous explanation generators for recommendation in group (A),
general lexically constrained generators in group (B), 
and our \our in group (C).}
\vspace{-2mm}
\scalebox{0.92}{
\setlength{\tabcolsep}{1mm}{
\begin{tabular}{cccccc}
\toprule
 Group & Methods      & \begin{tabular}[c]{@{}c@{}}Personalized \\ generation\end{tabular} & \begin{tabular}[c]{@{}c@{}}Aspect \\ planning\end{tabular} & \begin{tabular}[c]{@{}c@{}}Lexical\\ constraints\end{tabular} & \multicolumn{1}{c}{\begin{tabular}[c]{@{}c@{}}Random\\ keyphrases\end{tabular}} \\ \midrule
\multirow{3}{*}{(A)} &ExpansionNet~\cite{Ni2018PersonalizedRG} & \cmark   &  \cmark  &  \xmark   &     \xmark   \\
                     &Ref2Seq~\cite{Ni2019JustifyingRU}      & \cmark   &  \cmark  &  \xmark   &     \xmark   \\
                     &PETER~\cite{Li2021PersonalizedTF}        & \cmark   &  \cmark  &  \xmark   &     \xmark   \\ \midrule
\multirow{3}{*}{(B)} &NMSTG~\cite{Welleck2019NonMonotonicST}        & \xmark   &  \xmark  &  \cmark   &     \xmark   \\
                     &POINTER~\cite{Zhang2020POINTERCP}      & \xmark   &  \xmark  &  \cmark   &     \xmark   \\
                     &CBART~\cite{He2021ParallelRF}        & \xmark   &  \xmark  &  \cmark   &     \cmark   \\ \midrule
 (C)            &Ours        & \cmark   &  \cmark  &  \cmark   &     \cmark   \\ \bottomrule
\end{tabular}
}}
\vspace{-4mm}
\label{tab:comparison}
\end{table}

\section{Introduction}
\label{sec:intro}
Explaining, or justifying, recommendations in natural language is fast gaining traction in recent years~\cite{Li2021PersonalizedTF, Ni2018PersonalizedRG, lu2018like, SIGIR17-NRT, CIKM20-NETE, 2022-PEPLER, Ni2019JustifyingRU}, in order to show product information in a personalized style, and justify how the recommendation meets users' need. That is, given the pair of user and item, the system would generate an explanation such as ``\emph{nice TV with 4K display and Dolby Atmos!}''. 
To generate such high-quality personalized explanations which are coherent, relevant and informative,
recent studies introduce aspect planning, i.e.,~including different aspects~\cite{Li2021PersonalizedTF, Ni2018PersonalizedRG, 2022-PEPLER, Ni2019JustifyingRU} in the generation process so that the generated explanations will cover those aspects and thus be more relevant to products and to users' interests.

\begin{figure}[t]
    \centering
    \includegraphics[width=\linewidth]{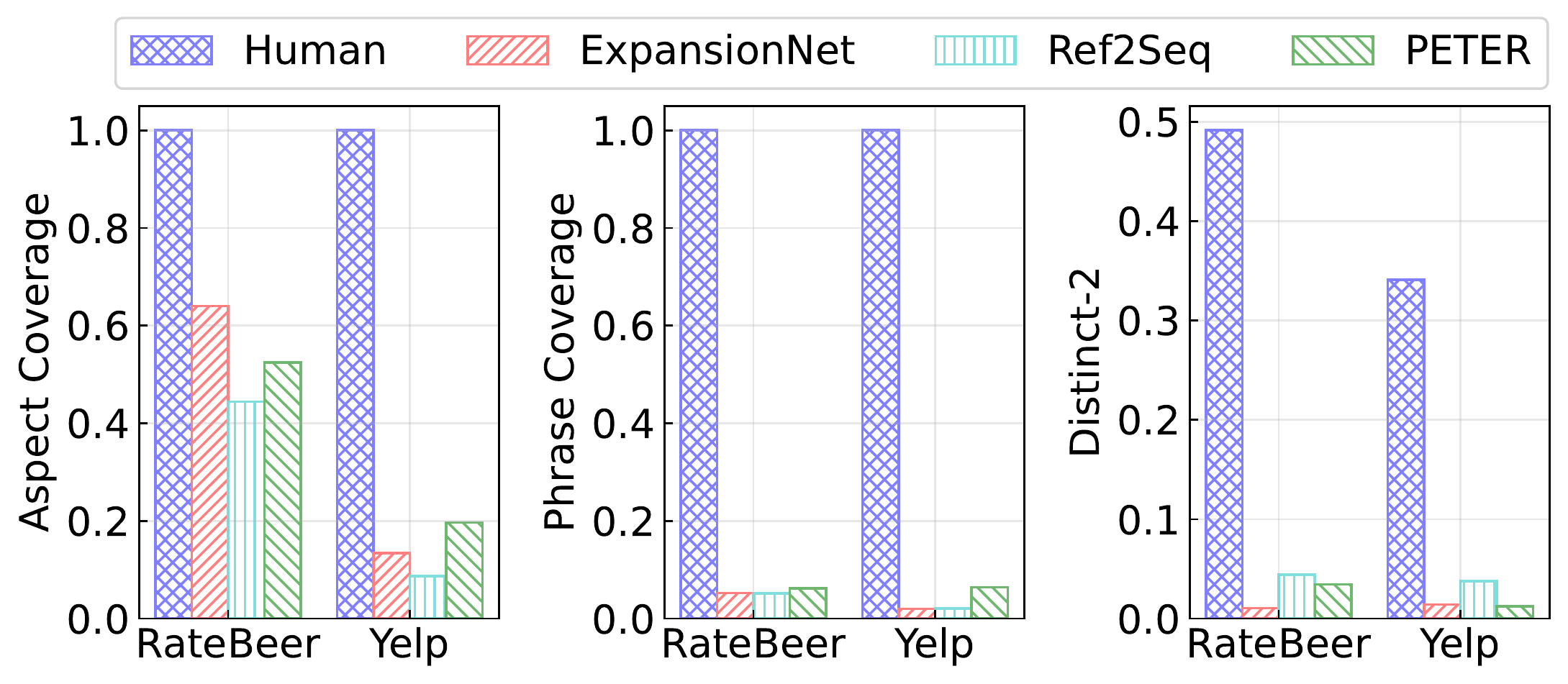}
    \caption{Preliminary experiments on the aspect coverage, phrase coverage, and Distinct-2 of generated explanations from previous models ExpansionNet~\cite{Ni2018PersonalizedRG}, Ref2Seq~\cite{Ni2019JustifyingRU} and PETER~\cite{Li2021PersonalizedTF} on RateBeer and Yelp datasets. Check details in~\Cref{app:motivation}}
    \label{fig:motivation}
    \vspace{-5mm}
\end{figure}

While promising, existing methods struggle to include accurate and highly specific information into explanations because aspects (e.g.,~\emph{screen} for a TV) mostly control the high-level sentiment or semantics of generated text (e.g.,~"\emph{good screen and audio}!"), but many informative product attributes are too specific to be accurately generated (e.g.,~"\emph{nice TV with 4K display and Dolby Atmos!}"). Although aspect-planning explanation generators try to harvest expressive and personalized natural-language-based explanations from users' textual reviews~\cite{Li2021PersonalizedTF, Ni2018PersonalizedRG, 2022-PEPLER, Ni2019JustifyingRU}, we observe that many informative and specific keyphrases in the training corpus (i.e.,~user reviews)  vanish in generated explanations according to our preliminary experiments. As \Cref{fig:motivation} shows,  generated explanations from previous methods miss many specific keyphrases and have much lower Distinct (diversity) scores than a human oracle. Hence, with aspects only, existing methods suffer from generating (1)~too general sentences (e.g.,~"\emph{good screen!}") that are hard to provide diverse and informative explanations to users; (2)~sentences with inaccurate details (e.g.,~"\emph{2K screen}" for a 4K TV), which are not relevant to the product and hurt users' trust.

To address the above problems, we propose to use more concrete constraints to recommendation explanations besides aspects. Specifically, we seek a model unifying \emph{lexical constraints} and \emph{aspect planning}. In this model, introducing lexical constraints guarantees the use of given keyphrases (e.g., "\emph{Dolby Atmos}") and thus includes specific and accurate information. Also, similar to the aspect selection of previous explanation generators~\cite{Ni2019JustifyingRU, Li2021PersonalizedTF}, such lexical constraints can come from multiple parties. For instance, \emph{explanation systems} select item attributes with some strategies; \emph{vendors} highlight product features; \emph{users} manipulate generated explanations by changing the lexical constraints of interest. Hence, the informativeness, relevance and diversity of generated explanations can be significantly improved compared to previous methods with aspect planning. Meanwhile, aspect planning remains useful when no specific given information but multiple aspects need to be covered. 

To achieve this goal of \underline{U}nifying aspect-planning and lexical \underline{C}onstraints for generating \underline{E}x\underline{p}lanat\underline{i}ons in Re\underline{c}ommendation, we present \our. There are some challenges of building \our. First, lexical constraints are incompatible with existing explanation generation models~(see group (A) in~\Cref{tab:comparison}), because they are mostly based on auto-regressive generation frameworks~\cite{Li2019GeneratingLA, Ni2018PersonalizedRG, Li2020KnowledgeEnhancedPR, Li2021KnowledgebasedRGF, Hua2019SentenceLevelCP, Moryossef2019StepbyStepSP} which cannot be guaranteed to contain lexical constraints in any positions with a ``left-to-right'' generation strategy. Second, although insertion-based generation models (see group (B) in~\Cref{tab:comparison}) are able to contain lexical constraints in generated sentences naturally, we find personalization or aspects cannot be simply incorporated with the "encoder-decoder" framework for existing insertion-based models. Existing tokens are strong signals for new tokens to be predicted, hence the model tends to generate similar sentences and ignore different references\footnote{In literature~\cite{Ni2019JustifyingRU}, the terminology \emph{references} refer to a user's personalized textual data such as the historical product reviews.} from encoders. 

For the first challenge, \our employs an insertion-based generation framework and conducts \emph{robust insertion pre-training} on a bi-directional transformer. During  robust pre-training, \our gains the basic ability to generate text and handle various lexical constraints. Specifically, inspired by Masked Language Modeling (MLM)~\cite{Devlin2019BERTPO}, we propose an insertion process that randomly inserts new tokens into sentences progressively so that \our is robust to random
lexical constraints. 
For the second challenge, \our uses \emph{personalized fine-tuning} for personalization and awareness of aspects. To tackle the issue of ``ignoring references'', we propose to view references as part of inserted tokens for the generator and hence the model learns to insert new tokens relevant to references. For aspect planning, we formulate aspects as a special insertion stage where aspect-related tokens will be first generated as a start for the following generation.
Finally, lexical constraints, aspect planning and personalized references are unified in the insertion-based generation framework.

Overall, \our is the first explanation generation model unifying aspect planning and lexical constraints. \our significantly improves \emph{relevance}, \emph{coherence} and \emph{informativeness} of generated explanations compared to existing methods. The main contributions of this paper are summarized as follows:
\begin{itemize}
    \item We show the limitations of only using  aspect planning in existing explanation generation, and propose to introduce lexical constraints for explanation generation.
    \item We present \our including robust insertion pre-training and personalized fine-tuning to unify aspect planning, lexical constraints and references in an insertion-based generation framework.
    \item We conduct extensive experiments on two datasets. Objective metrics and human evaluations show that \our can largely improve the diversity, relevance, coherence and informativeness of generated explanations.
\end{itemize}

\begin{figure*}
    \centering
    \includegraphics[width=0.95\linewidth]{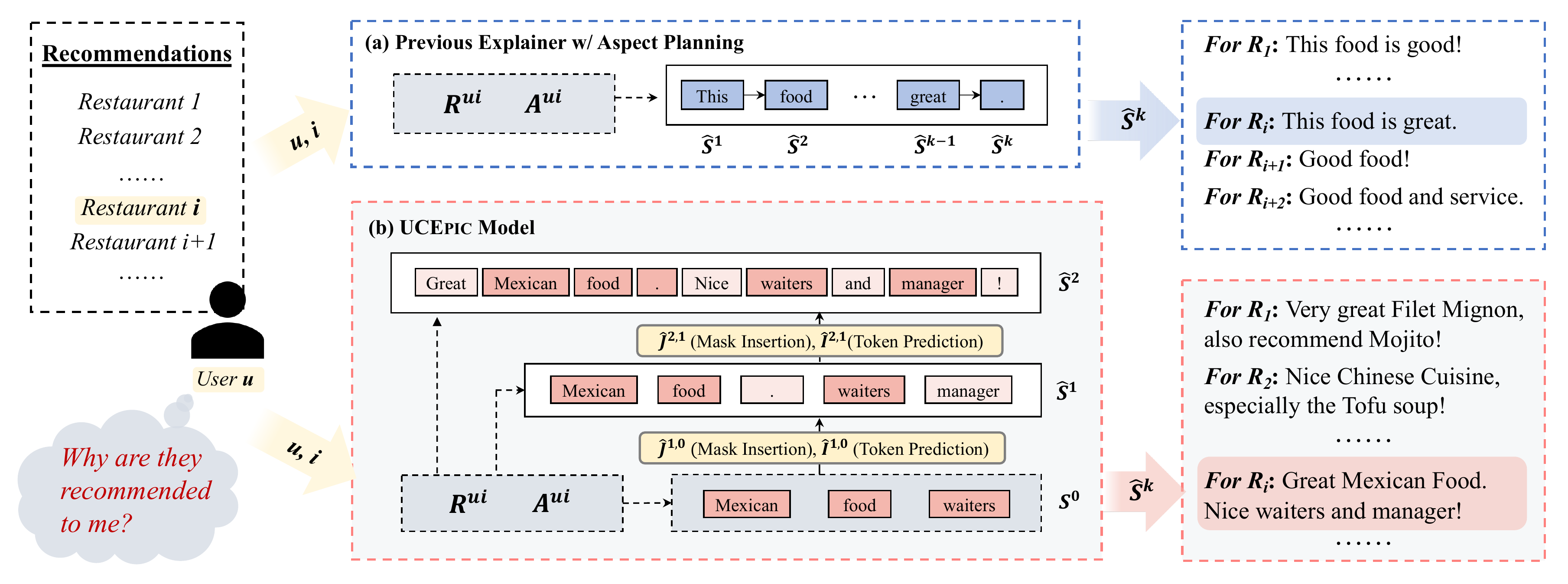}
    \vspace{-4mm}
    \caption{Overview of generating explanations for a given user and recommended items using (a) an aspect-planning autoregressive generation model; using (b) our \our that unifies aspect-planning and lexical constraints.}
    \label{fig:ucepic}
    \vspace{-2mm}
\end{figure*}

%% file: 2_related.tex
\section{Related Work}
\xhdr{Explanation Generation For Recommendation.}
Generating explanations of recommended items for users has been studied for a long time with various output formats~\cite{zhang2020explainable, zhang2014explicit, gao2019explainable} (e.g.,~item aspects, attributes, similar users).  Recently, natural-language-based explanation generation has drawn great attention~\cite{Li2021PersonalizedTF, Ni2018PersonalizedRG, lu2018like, SIGIR17-NRT, CIKM20-NETE, 2022-PEPLER, Ni2019JustifyingRU} to generate post-hoc explanations or justifications in personalized style. For example, \citet{SIGIR17-NRT} applied RNN-based model to generate explanations based on predicted ratings. To better control the explanation generation process, \citet{Ni2019JustifyingRU} extracted aspects and controlled the semantics of generated explanations conditioned on different aspects, and \citet{Li2021PersonalizedTF} proposed a personalized transformer model to generate explanations based on given item features.   Also, the review generation area is highly related since explanation generation methods usually harvest expressive and informative explanations from user reviews. Many controllable review generators~\cite{Tang2016ContextawareNL, Zhou2017LearningTG, Ni2018PersonalizedRG} are tailored to explanation generations as baseline models in early experiments. 
Although previous works continued increasing the controllability of generation, they are all on the basis of auto-regressive generation frameworks~\cite{Li2019GeneratingLA, Ni2018PersonalizedRG, Li2020KnowledgeEnhancedPR, Li2021KnowledgebasedRGF, Hua2019SentenceLevelCP, Moryossef2019StepbyStepSP} thus only considering aspect planning. In our work, \our increases the controllability, informativeness of generated explanations by unifying aspect planning and lexical constraints under an insertion-based generation framework.

\xhdr{Lexically Constrained Text Generation.}
Lexically constrained generation requires that generated text contain the lexical constraints (e.g.,~keywords). Early works usually involve special decoding methods. \citet{Hokamp2017LexicallyCD} proposed a lexical-constrained grid beam search decoding algorithm to incorporate constraints. \citet{Post2018FastLC} presented an algorithm for lexically constrained decoding with reduced complexity in the number of constraints. \citet{Hu2019ImprovedLC} further improved decoding by a vectorized dynamic beam allocation. \citet{Miao2019CGMHCS} introduced a sampling-based conditional decoding method, where the constraints are first placed in a template, then decoded words under a Metropolis-Hastings sampling. Special decoding methods usually need a high running time complexity. Recently, \citet{Zhang2020POINTERCP} implemented hard-constrained generation with  $\mathcal{O}(\log n)$ time complexity by language model pre-training and insertion-based generation~\cite{Stern2019InsertionTF, Gu2019LevenshteinT, Chan2019KERMITGI, Gu2019InsertionbasedDW} used in machine translation. CBART~\cite{He2021ParallelRF} uses the pre-trained model BART~\cite{Lewis2020BARTDS} and the encoder and decoder are used for 
instructing insertion
and 
predicting mask respectively.

%% file: 3_method.tex
\section{Methodology}

\label{sec:method}
We describe aspect planning and lexical constraints for explanation generation as follows. Given a user persona $\mathrm{R}^u$, item profile $\mathrm{R}^i$ for user $u$ and item $i$ as references,
the generation model under aspect planning outputs the explanation $\mathrm{E}^{ui}$ related to an aspect $\mathrm{A}^{ui}$ but not necessarily including
some specific words. 
Whereas for lexical constraints, given several lexical constraints (e.g,~phrases or keywords) $\mathrm{C}^{ui} = \{c_1, c_2,\dots,c_m\}$, the model will generate an explanation $\mathrm{E}^{ui} = (w_1,w_2,\dots,w_n)$ that has to exactly include all given lexical constraints $c_i$, which means $c_i = (w_j,\dots, w_k)$. The lexical constraints can be from users, businesses, or item attributes recommended by personalized systems in a real application. \our unifies the two kinds of constraints in one model~\footnote{\our has two modes: generating under aspect planning or generating under lexical constraints}. We study only the explanation generation method and assume aspects and lexical constraints are given. Our notations are summarized in~\Cref{tab:notation}

\begin{table}[t]
\centering
\caption{Notation.}
\vspace{-3mm}
\label{tab:notation}
\begin{tabular}{lp{6cm}}
\toprule
\textbf{Notation} & \textbf{Description} \\ \midrule
$\mathrm{R}^u$, $\mathrm{R}^i$ & historical review profile of user $u$ and item $i$. \\ 
$\mathrm{E}^{ui}$ & generated explanation when item $i$ is recommended to user $u$. \\ 
$\mathrm{A}^{ui}$ & aspects controlling explanation generation for item $i$ and user $u$.\\ 
$\mathrm{C}^{ui}$ & lexical constraints (e.g., keywords) controlling explanation generation for item $i$ and user $u$.\\ 
$S^{k}$, $\hat{S}^{k}$ & text sequence of the $k$-th stage generation. $S^{k}$ is training data and $\hat{S}^{k}$ is model prediction.\\ 
$I^{k,k-1}$, $\hat{I}^{k,k-1}$& intermediate sequence between $S^{k-1}$ and $S^{k}$. (training data and model prediction)\\ 
$J^{k, k-1}$, $\hat{J}^{k,k-1}$& insertion number sequence between $S^{k-1}$ and $S^{k}$. (training data and model prediction)\\
$\mathbf{D}$& a bi-directional transformer for encoding.\\
$\mathbf{H}_\mathit{MI}$ & a linear projection layer for insertion numbers. \\ 
$\mathbf{H}_\mathit{TP}$ & a multilayer perceptron with activation function for token prediction. \\ \bottomrule
\end{tabular}
\vspace{-4mm}
\end{table}

\subsection{Robust Insertion}
\subsubsection{Motivation} Previous explanation generation methods~\cite{Ni2019JustifyingRU, Li2021PersonalizedTF} generally adopt auto-regressive generation conditioned on some personalized inputs (e.g.,~personalized references and aspects). As shown in~\Cref{fig:ucepic} (a), the auto-regressive process generates words in a ``left-to-right'' direction so lexical constraints are difficult to be contained in the generation process. However, for the insertion-based generation in~\Cref{fig:ucepic} (b) which progressively inserts new tokens based on existing words, lexical constraints can be easily contained by viewing constraints as a starting stage of insertion. 


\subsubsection{Formulation.}
The insertion-based generation can be formulated as a progressive sequence of $K$ stages $S = \{S^0, S^1,\dots,S^{K-1},S^{K}\}$, where $S^0$ is the stage of lexical constraints and $S^{K}$ is our final generated text. For $k\in \{1,\dots, K\}$, $S^{k-1}$ is a sub-sequence of $S^k$. The generation procedure finishes when \our does not insert any new tokens into $S^{K}$. In the training process, all sentences are prepared as training pairs. Specifically, pairs of text sequences are constructed at adjacent stages $(S^{k-1}, S^{k})$ that reverse the insertion-based generation process. Each explanation $E^{ui}$ in the training data is broken into a consecutive series of pairs:
$(S^{0}, S^{1}), (S^{1}, S^{2}), \dots, (S^{K-1}, S^{K})$, and when we construct the training data, the final stage $S^K$ is our explanation text $E^{ui}$.

\begin{table}[t]
\vspace{-3mm}
\caption{Data construction examples.}
\vspace{-3mm}
\label{tab:data_cons}
\small
\begin{tabular}{@{}lp{5.5cm}@{}}
\toprule
\small
\textbf{Data}         & \textbf{Example}                                                                                                                     \\ \midrule
$S^K$ (sentence)            & \textless{}s\textgreater{}Good tacos. Love the crispy citrus + tropical fruits flavor. \textless{}/s\textgreater{}               \\
 $I^{K, K-1}$                     & \textless{}s\textgreater{}\texttt{[MASK]} tacos. Love the \texttt{[MASK]} \texttt{[MASK]} + tropical fruits flavor. \textless{}/s\textgreater{} \\
$J^{K, K-1}$                    & \texttt{[1 0 0 0 2 0 0 0 0 0 0]}                                                                                                        \\
$S^{K-1}$            & \textless{}s\textgreater{}tacos. Love the + tropical fruits flavor. \textless{}/s\textgreater{}                                  \\
$\dots$                   & $\dots$                                                                                                                                  \\
$S^0$ (lexical constraints) & \textless{}s\textgreater{}tropical fruits flavor \textless{}/s\textgreater{}                                                         \\ \bottomrule
\end{tabular}
\end{table}

\subsubsection{Data Construction.}
Given a sequence stage $S^k$, we obtain the previous stage $S^{k-1}$ by two operations, masking and deletion. Specifically, we randomly mask the tokens in a sequence by probability $\mathrm{p}$ as MLM to get the intermediate sequence $I^{k,k-1}$. Then, \texttt{[MASK]} tokens are deleted from the intermediate sequence $I^{k, k-1}$ to obtain the stage $S^{k-1}$. The numbers of deleted \texttt{[MASK]} tokens after each token in $I^{k, k-1}$ are recorded as an insertion number sequence $J^{k, k-1}$. Finally, each training instance contains four sequences $(S^{k-1}, I^{k, k-1}, J^{k, k-1}, S^{k})$. We include a simple example for the data construction process in~\Cref{tab:data_cons}. Since we delete $T*\mathrm{p}$ tokens in sequence $S^{k}$ where $T$ is the length of $S^{k}$, the average number of $K$ is $\log_{\frac{1}{\mathrm{1-p}}}T$. Models trained on this data will easily re-use the knowledge from BERT-like models which have a similar pre-training process of masked word prediction.

Insertion generation (see~\Cref{alg:Insertion}) is an inverse process of data construction. For each insertion step prediction from $\hat{S}^{k-1}$ to $\hat{S}^{k}$, the model will recover text sequences by two operations, mask insertion and token prediction. In particular, \our first inserts \texttt{[MASK]} tokens between any two existing tokens in $\hat{S}^{k-1}$ to get $\hat{I}^{k, k-1}$ according to $\hat{J}^{k, k-1}$ predicted by an insertion prediction head. Then, \our with a language modeling head predicts the masked tokens in $\hat{I}^{k, k-1}$, and recovers \texttt{[MASK]} tokens into words to obtain the $\hat{S}^{k}$.

\begin{algorithm}[t]
\caption{\textbf{Insertion in the $k$-th Stage}}\label{alg:Insertion}
    \begin{algorithmic}
    \Procedure{Insertion}{$\hat{S}^{k-1}$} 
        \State $\hat{J}^{k,k-1} \gets$ predict number of masks from $\hat{S}^{k-1}$ via~\cref{eq:1} ;
        \State $\hat{I}^{k,k-1} \gets$ build intermediate sequence from $\hat{J}^{k,k-1}$ and $\hat{S}^{k-1}$;
        \State $\hat{S}^{k} \gets$ predict masked tokens in $\hat{I}^{k,k-1}$ via~\cref{eq:2};
        \State \Return predicted sequence $\hat{S}^{k}$;
    \EndProcedure
    \end{algorithmic}
\end{algorithm}

\subsubsection{Modules}
\our uses a bi-directional Transformer architecture with two different prediction heads for mask insertion and token prediction. The architecture of the model is closely related to that used in RoBERTa~\cite{Liu2019RoBERTaAR}. The bi-directional Transformer $\mathbf{D}$ will predict the mask insertion numbers and word tokens with two heads $\mathbf{H}_\mathit{MI}$ and $\mathbf{H}_\mathit{TP}$ respectively. $\mathbf{H}_\mathit{TP}$ is a multilayer perceptron (MLP) with activation function GeLU~\cite{Hendrycks2016GaussianEL} and $\mathbf{H}_\mathit{MI}$ is a linear projection layer. Finally, our predictions of mask insertion numbers and word tokens are computed as:
\begin{align}
    \label{eq:1}
    y_\mathit{MI} &= \mathbf{H}_\mathit{MI}(\mathbf{D}(\hat{S}^{k-1})),\ \hat{J}^{k,k-1} = \mathrm{argmax}(y_\mathit{MI}) \\
    \label{eq:2}
    y_\mathit{TP} &= \mathbf{H}_\mathit{TP}(\mathbf{D}(\hat{I}^{k, k-1})),\ \hat{S}^{k} = \mathrm{argmax}(y_\mathit{TP})
\end{align}
where $y_\mathit{MI} \in \mathbb{R}^{l_s \times d_\mathit{ins}}$ and $y_\mathit{TP} \in \mathbb{R}^{l_I \times d_\mathit{vocab}}$, $l_s$ and $l_I$ are the length of $\hat{S}^{k-1}$ and $\hat{I}^{k, k-1}$ respectively, $d_\mathit{ins}$ is the maximum number of insertions and $d_{\mathit{vocab}}$ is the size of vocabulary. $\hat{I}^{k, k-1}$ is obtained by inserting \texttt{[MASK]} tokens into $\hat{S}^{k-1}$ according to $\hat{J}^{k,k-1}$.

Because the random insertion process is more complicated to learn than the traditional autoregressive generation process, we first pre-train \our with our robust insertion method for general text generation without personalization. The pre-trained model can generate sentences from randomly given lexical constraints.

\subsection{Personalized References and Aspect Planning}
\subsubsection{Motivation} To incorporate personalized references and aspects, one direct method is to have another text and aspect encoder and insertion generation conditioned on the encoder like the sequence-to-sequence model~\cite{Sutskever2014SequenceTS}. However, we find the pre-trained insertion model with another encoder will generate similar sentences with different personalized references and aspects. The reason is the pre-trained insertion model views the lexical constraints or existing tokens in text sequences as a strong signal to determine new inserted tokens. Even if our encoder provides personalized features, the model tends to overfit features from existing tokens. Without lexical tokens providing different starting stages, generated sentences are usually the same.

\subsubsection{Formulation}
To better learn personalization, we propose to view references and aspects as special existing tokens during the insertion process. Specifically, we construct a training stage $S^k_{+}$ to include references and aspects as:
\begin{equation}
\begin{split}
    S^k_{+} &= [\mathrm{R}^{ui}, \mathrm{A}^{ui}, S^k] \\
    &=[w^r_0,\dots,w^r_{|\mathrm{R}^{ui}|}, w^a_0,\dots,w^a_{|\mathrm{A}^{ui}|}, w_0,\dots,w_{|S^k|}]
\end{split}
\end{equation}
where $\mathrm{R}^{ui}$, $\mathrm{A}^{ui}$ denote personalized references and aspects; $w^r$, $w^a$ and $w$ are tokens or aspect ids in references, aspects and insertion stage tokens respectively. Because insertion-based generation relies on token positions to insert new tokens, we create token position ids in Transformer starting from 0 for $\mathrm{R}^{ui}$, $\mathrm{A}^{ui}$ and $S^k$ respectively in order to make it consistent for $S^k$ between pre-training and fine-tuning. Similarly, we obtain the insertion number sequence $J^{k, k-1}_{+} = [\mathbf{0}_{|\mathrm{R}^{ui}|}, \mathbf{0}_{|\mathrm{A}^{ui}|}, J^{k,k-1}]$ and intermediate training stage $I^{k,k-1}_{+} = [\mathrm{R}^{ui}, \mathrm{A}^{ui}, I^{k,k-1}]$, where $\mathbf{0}_{|\mathrm{R}^{ui}|}$ and $\mathbf{0}_{|\mathrm{A}^{ui}|}$ are zero vectors which have same length as $\mathrm{R}^{ui}$ and $\mathrm{A}^{ui}$ respectively, because we do not insertion any tokens into references and aspects.

\subsubsection{Modules}
We encode $\hat{S}^k_{+}$ and $\hat{I}_{+}^{k,k-1}$ with bi-directional Transformer $\mathbf{D}$ to get the insertion numbers $y_{MI}$ and predicted tokens $y_{TP}$ as follows:
\begin{align}
[\mathrm{O}_S^{R^{ui}},\mathrm{O}_S^{A^{ui}}, \mathrm{O}^{S^{k}}] &= \mathbf{D}(\hat{S}^k_{+}) \\
[\mathrm{O}_I^{R^{ui}},\mathrm{O}_I^{A^{ui}}, \mathrm{O}^{I^{k,k-1}}] &= \mathbf{D}(\hat{I}_{+}^{k,k-1}) \\
y_\mathit{MI} &= \mathbf{H}_\mathit{MI}(\mathrm{O}^{S^{k}})\\
y_\mathit{TP} &= \mathbf{H}_\mathit{TP}(\mathrm{O}^{I^{k,k-1}})
\end{align}
Similar as \Cref{eq:1} and \Cref{eq:2}, we can get $\hat{J}^{k, k-1}$ and $\hat{S}^k$ by $\mathrm{argmax}$ operation.
Because personalized references and aspects are viewed as special existing tokens, \our will directly incorporate token-level information as generation conditions and hence generates diverse explanations.

Recall that existing text sequences are strong signals for token prediction. For better aspect-planning generation, we design two starting stages $S^0_{+a}$ and $S^0_{+l}$ for aspects and lexical constraints respectively. In particular, we expect the aspect-related tokens can be generated at the starting stage (i.e.,~no existing tokens) according to given aspects and personalized references. Hence, the aspect starting stage is $S^0_{+a}=[\mathrm{R}^{ui}, \mathrm{A}^{ui}]$. Lexical constraint starting stage is $S^0_{+l} = [\mathrm{R}^{ui}, \mathrm{A}^{\mathit{pad}}, \mathrm{C}^{ui}]$ where $\mathrm{A}^{\mathit{pad}}$ is a special aspect that is used for lexical constraints. During training, we sample $S^0_{+a}$ with probability $p$ to ensure \our learns aspect-related generation effectively which is absent in pre-training.


\subsection{Model Training}
The training process of \our is to learn the inverse process of data generation. Given stage pairs $(S_{+}^{k-1}, S_{+}^k)$ and training instance $(S_{+}^{k-1}, I_{+}^{k, k-1}, J_{+}^{k, k-1}, S_{+}^{k})$ from pre-processing~\footnote{For fine-tuning with personalized references and aspects, we train the model with stage pairs $(S_{+}^{k-1}, S_{+}^k)$ and training instance $(S_{+}^{k-1}, I_{+}^{k, k-1}, J_{+}^{k, k-1}, S_{+}^{k})$}, we optimize the following objective:
\begin{equation}
    \begin{split}
        \mathcal{L}&= -\log p(S^k_{+}|S^{k-1}_{+}) \\
        &= -\log \underbrace{p(S^k_{+}, J^{k, k-1}_{+}|S^{k-1}_{+})}_{\text{Unique $J$ assumption}} \\
        &= -\log p(S^k_{+}|J^{k,k-1}_{+}, S^{k-1}_{+})p(J^{k,k-1}_{+}|S^{k-1}_{+}) \\
        &= -\log\underbrace{p(S^k_{+}|I^{k, k-1}_{+})}_{\text{Token prediction}}\underbrace{p(J^{k,k-1}_{+}|S^{k-1}_{+})}_{\text{Mask insertion}}, \\
        &\text{where}\ I^{k, k-1}_{+}=\mathrm{MaskInsert}(J^{k,k-1}_{+}, S^{k-1}_{+})
    \end{split}
\label{eq:obj}
\end{equation}
where $\mathrm{MaskInsert}$ denotes the mask token insertion. 
We make a reasonable assumption that \emph{$J^{k, k-1}_{+}$ is unique given $(S^k_{+}, S^{k-1}_{+})$}. This assumption is usually true unless in some corner cases multiple $J^{k,k-1}_{+}$ could be legal (e.g., masking one ``moving'' word in ``a moving moving moving van''); $I^{k, k-1}_{+}$ by definition is the intermediate sequence, which is equivalent to the given $(J^{k,k-1}_{+}, S^{k-1}_{+})$.
In~\Cref{eq:obj}, we jointly learn
\begin{inparaenum}[(1)]
\item likelihood of mask insertion number for each token from \our with $\mathbf{H}_\mathit{MI}$, and
\item likelihood of word tokens for the masked tokens from \our with $\mathbf{H}_\mathit{TP}$.
\end{inparaenum}

Similar to training BERT~\cite{Devlin2019BERTPO}, we optimize only the masked tokens in token prediction. The selected tokens to mask have the probability $0.1$ to stay unchanged and the probability $0.1$ to be randomly replaced by another token in the vocabulary. For mask insertion number prediction, most numbers in $J^{k, k-1}_{+}$ are $0$ because we do not insert any tokens between the existing two tokens in most cases. To balance the insertion number, we randomly mask the $0$ in $J^{k, k-1}_{+}$ with probability $\mathrm{q}$. Because our mask prediction task is similar to masked language models, the pre-trained weights from RoBERTa~\cite{Liu2019RoBERTaAR} can be naturally used for initialization of \our to obtain prior knowledge.

\subsection{Inference}

At inference time, we start from the given aspects $\mathrm{A}^{ui}$ or lexical constraint $\mathrm{C}^{ui}$ to construct starting stage $S^0_{+a}$ or $S^0_{+l}$ respectively. Then, \our predicts $\{\hat{S}^{1}_{+},\dots, \hat{S}^{K}_{+}\}$ repeatedly until no additional tokens generated or reaching the maximum stage number. We obtain final generated explanation $\hat{S}^K$ from $\hat{S}^K_{+}$ by removing $\mathrm{R}^{ui}$ and $\mathrm{A}^{ui}$. Without loss of generality, we show the inference details from $\hat{S}^{k-1}_{+}$ stage to $\hat{S}^k_{+}$ stage: (1)~given $\hat{S}^{k-1}_{+}$ \our uses $\mathbf{H}_\mathit{MI}$ to predict $\hat{J}^{k, k-1}_{+}$ insertion number sequence\footnote{We set predicted insertion number as 0 for given phrases in $S^0_{+l}$, to prevent given phrases from modification.}; (2)~given $\hat{I}^{k, k-1}_{+}$ from $\mathrm{MaskInsert}(\hat{J}^{k, k-1}_{+}, \hat{S}^{k-1}_{+})$, \our can use $\mathbf{H}_{TP}$ to predict $\hat{S}^{k}_{+}$ with a specific decoding strategy (e.g.,~greedy search, top-K sampling). (3)~given $\hat{S}^{k}_{+}$, \our meets the termination requirements or executes step (1) again. The termination criterion can be a maximum interation number or \our does not insert new tokens into $\hat{S}^{k}_{+}$.

%% file: 4_experiments.tex
\section{Experiments}
\label{sec:exp}

\begin{table}[t]
\centering
\caption{Statistics of datasets}
\vspace{-2mm}
\scalebox{1}{
\setlength{\tabcolsep}{1mm}{
\begin{tabular}{lcccccc}
\toprule
\textbf{Dataset}   & \textbf{Train}   & \textbf{Dev}    & \textbf{Test}   & \textbf{\#Users} & \textbf{\#Items} & \textbf{\#Aspects} \\ \midrule
RateBeer  & 16,839  & 1,473  & 912    & 4,385    & 6,183  &  8 \\
Yelp      & 252,087 & 37,662 & 12,426 & 235,794  & 22,412  &  59\\\bottomrule
\end{tabular}
}
}
\vspace{-5mm}
\label{tab:data}
\end{table}

\begin{table*}[]
\centering
\caption{Performance comparison of the explanation generation models (ExpansionNet, Ref2Seq, PETER), lexically constrained generation models (NMSTG, POINTER, CBART) and \our. All values are in percentage (\%). We \underline{underline} the highest scores of aspect-planning generation results and the highest scores of lexically constrained generation are \textbf{bold}.}
\vspace{-2mm}
\begin{tabular}{lcccccccccccccc}
\toprule
     & \multicolumn{7}{c}{\textbf{RateBeer}}                                                                                                                                                           & \multicolumn{7}{c}{\textbf{Yelp}}                                                                                                                                                      \\ \cmidrule(l){2-8} \cmidrule(l){9-15}
\multicolumn{1}{c}{\textbf{Models}}      & B-1                     & B-2                     & D-1                              & D-2                      & M                       & R                        & BS                       & B-1                     & B-2                     & D-1                     & D-2                      & M                       & R                        & BS                       \\ \midrule
\multicolumn{1}{l|}{Human-Oracle} & --                      & --                      & 8.30                              & 49.16                     & --                      & --                       & --                       & --                      & --                      & 3.8                     & 34.1                     & --                      & --                       & --                       \\ \midrule
\multicolumn{15}{c}{\textit{Aspect-planning generation}}                                                                                                                                                                                                                                                                                                                                                                     \\ \midrule
\multicolumn{1}{l|}{ExpansionNet} & 8.96           & 1.79          & 0.20          & 1.05           & 16.30          & 10.13          & 75.58          & 4.92           & 0.47          & 0.18          & 1.40           & 7.78           & 5.42           & 76.27          \\
\multicolumn{1}{l|}{Ref2Seq}      & 17.15          & 4.17          & 0.95          & 4.41           & 16.66          & 15.66          & 80.76          & 8.34           & 0.98          & 0.46          & 3.77           & 7.58           & 11.19          & 82.66          \\
\multicolumn{1}{l|}{PETER}        & 25.25                    & \underline{5.35}            & 0.74                              & 3.44                      & 19.19                    & \underline{20.34}            & \underline{84.03}            & \underline{14.26}           & \underline{2.25}            & 0.26                     & 1.23                      & \underline{12.25}           & \underline{14.75}            & 82.55                     \\
\multicolumn{1}{l|}{\our}       & \underline{27.42}           & 2.89                     & \underline{4.49}                     & \underline{29.23}            & \underline{19.54}           & 15.48                    & 83.53                     & 8.03                     & 0.72                     & \underline{1.89}            & \underline{14.75}            & 8.10                     & 11.58                     & \underline{83.53}            \\ \midrule
\multicolumn{15}{c}{\textit{Lexically constrained generation}}                                                                                                                                                                                                                                                                                                                                                               \\ \midrule
\multicolumn{1}{l|}{ExpansionNet} & 5.41                     & 0.49                     & 0.97                              & 4.91                      & 6.09                     & 5.55                      & 76.14                     & 1.49                     & 0.08                     & 0.40                     & 1.90                      & 2.19                     & 1.93                      & 73.68                    \\
\multicolumn{1}{l|}{Ref2Seq}      & 17.94          & 4.50          & 1.09          & 5.49           & 17.03          & 15.17          & 83.72          & 6.38           & 0.77          & 0.51          & 3.64           & 7.02           & 10.58          & 82.88          \\
\multicolumn{1}{l|}{PETER}        & 15.03                    & 2.46                     & 2.04                              & 11.40                     & 9.49                     & 13.27                     & 79.08                     & 7.59                     & 1.32                     & 1.52                     & 8.70                      & 7.64                     & 12.24                     & 80.89                     \\
\multicolumn{1}{l|}{NMSTG}        & 22.82                    & 2.30                     & 6.02                              & 50.39          & 15.17                    & 15.35                     & 82.31                     & 13.67                    & 0.77                     & \textbf{4.57}            & \textbf{57.02}                     & 9.64                     & 11.13                     & 80.80                     \\
\multicolumn{1}{l|}{POINTER}      & 6.00                     & 0.31                     & \textbf{11.24}                             & \textbf{56.02}                     & 7.41                     & 11.21                     & 81.80                     & 1.50                     & 0.06                     & 5.49                     & 29.76           & 3.24                     & 5.23                      & 80.85                     \\
\multicolumn{1}{l|}{CBART}        & 2.49 & 0.54 & 8.49 & 34.74 & 8.45 & 13.84 & 83.30 & 2.19 & 0.60 & 5.32 & 26.79 & 9.41 & 15.00 & 84.08 \\
\multicolumn{1}{l|}{\our}       & \textbf{27.97}           & \textbf{5.09}            & 5.24                              & 32.04                     & \textbf{19.90}           & \textbf{17.05}            & \textbf{84.03}            & \textbf{13.77}           & \textbf{3.06}            & 2.85                     & 20.39                     & \textbf{14.45}           & \textbf{16.92}            & \textbf{84.55}            \\ \bottomrule
\end{tabular}
\label{tab:main_perform}
\vspace{-2mm}
\end{table*}

\subsection{Datasets}
For \emph{pre-training}, we use English Wikipedia\footnote{https://dumps.wikimedia.org/} for robust insertion training which has $11.6$ million sentences. For fair comparison with baselines pre-trained on a general corpus, we use Wikipedia as the pre-training dataset; and for \emph{fine-tuning}, we use Yelp\footnote{\url{https://www.yelp.com/dataset}} and RateBeer~\cite{McAuley2013FromAT} to evaluate our model (see~\Cref{tab:data}). We further filter the reviews with a length larger than $64$. For each user, following~\citet{Ni2019JustifyingRU}, we randomly hold out two samples from all of their reviews to construct the development and test sets. Following previous works~\cite{Ni2019JustifyingRU, Ni2017EstimatingRA}, we employ an unsupervised aspect extraction tool~\cite{Li2022UCTopicUC} to obtain phrases and corresponding aspects for lexical constraints and aspect planning respectively. The number of aspects for each dataset is determined by the tool automatically and aspects provide coarse-grained semantics of generated explanations. Note that, typically, the number of aspects is much smaller than the number of lexical constraints and aspects are more high-level.

\subsection{Baselines} 
We consider two groups of baselines for automatic evaluation to evaluate  model effectiveness.
The first group is existing text generation models for recommendation with \emph{aspect planning}.
\begin{itemize}
\item \textbf{ExpansionNet}~\cite{Ni2018PersonalizedRG}, generates reviews conditioned on different aspects extracted from a given review title or summary.
\item \textbf{Ref2Seq}~\cite{Ni2019JustifyingRU}, a Seq2Seq model incorporates contextual information from reviews and uses fine-grained aspects to control explanation generation.
\item \textbf{PETER}~\cite{Li2021PersonalizedTF}, a Transformer-based model that uses user- and item-IDs and given phrases to predict the words in target explanation generation. This baseline can be considered as a state-of-the-art model for explainable recommendation.
\end{itemize}
We compare those baselines under both aspect planning and lexical constraints. Specifically, we feed lexical constraints (i.e.,~keyphrases) into models and expect models copy keyphrases to generated text.

The second group includes general natural language generation models with \emph{lexical constraints}:
\begin{itemize}
\item \textbf{NMSTG}~\cite{Welleck2019NonMonotonicST}, a tree-based text generation scheme that from given lexical constraints in prefix tree form, the model generates words to its left and right, yielding a binary tree. 
\item \textbf{POINTER}~\cite{Zhang2020POINTERCP}, an insertion-based generation method pre-trained on constructed data based on dynamic programming. 
\item \textbf{CBART}~\cite{He2021ParallelRF}, uses the pre-trained BART~\cite{Lewis2020BARTDS} and instructs the decoder to insert and replace tokens by the encoder.
\end{itemize}
The second group of baselines cannot incorporate aspects or personalized information as references. These models are trained and generate text solely based on given lexical constraints. We do not include explanation generation methods such as NRT~\cite{Chen2017ExplainingRB}, Att2Seq~\cite{Zhou2017LearningTG} and ReXPlug~\cite{hada2021rexplug}, non-natural-language explainable recommenders such as EFM~\cite{zhang2014explicit} and DEAML~\cite{gao2019explainable}, and lexically constrained methods CGMH~\cite{Miao2019CGMHCS}, GBS~\cite{Hokamp2017LexicallyCD} because PETER and CBART reported better performance than these models. We also had experiments with "encoder-decoder" based \our as mentioned in~\Cref{sec:intro}, but this model generates same sentences for all user-item pairs hence we do not include it as a baseline.
Detailed settings of baselines can be found in~\Cref{app:baseline}.

\subsection{Evaluation Metrics}
\label{sec:metric}
We evaluate the generated sentences from two aspects: generation quality and diversity.
Following~\citet{Ni2019JustifyingRU, Zhang2020POINTERCP}, we use n-gram metrics including BLEU (B-1 and B-2)~\cite{Papineni2002BleuAM}, METEOR (M)~\cite{Banerjee2005METEORAA} and ROUGE-L (R-L)~\cite{Lin2004ROUGEAP} which measure the similarity between the generated text and human oracle. As for generation diversity, we use Distinct (D-1 and D-2)~\cite{Li2016ADO}. We also introduce BERT-score (BS)~\cite{Zhang2019BERTScoreET} as a semantic rather than n-gram metrics.

\subsection{Implementation Details}
We use \texttt{RoBERT-base}~\cite{Liu2019RoBERTaAR} (\#params $\approx$ 130M, other pretrained model sizes are listed in~\Cref{app:baseline}). In training data construction, we randomly mask $\mathrm{p}=0.2$ tokens in $S^k$ to obtain $I^{k, k-1}$. $0$ in $J^{k, k-1}$ are masked by probability $\mathrm{q}=0.9$. The tokenizer is byte-level BPE following RoBERTa. For \emph{pre-training}, the learning rate is 5e-5, batch size is $512$ and our model is optimized by AdamW~\cite{loshchilov2018decoupled} in $1$ epoch. For \emph{fine-tuning} on downstream tasks, the learning rate is 3e-5, and the batch size is $128$ with the same optimizer as pre-training. The training epoch is $10$ and we select the best model on the development set as our final model which is evaluated on test data. We randomly sample one aspect and one phrase from the target text as the aspect for planning and lexical constraint respectively~\footnote{More details are in~\url{https://github.com/JiachengLi1995/UCEpic}. We also released an additional checkpoint pre-trained on personalized review datasets including Amazon Reviews~\cite{Ni2019JustifyingRU} and Google Local~\cite{yan23showcases}.}.

\subsection{Automatic Evaluation}
\subsubsection{Overall Performance}
In \Cref{tab:main_perform}, we report evaluation results for different generation methods. For aspect-planning generation, \our can achieve comparable results as the state-of-the-art model PETER. Specifically, although PETER obtains better B-2 and ROUGE-L than our model, the results from \our are significantly more diverse than PETER. A possible reason is that auto-regressive generation models such as PETER tend to generate text with higher n-gram metric results than insertion-based generation models, because auto-regressive models generate a new token solely based on left tokens while (insertion-based) \our considers tokens in both directions. Despite the intrinsic difference, \our still achieves comparable B-1, Meteor and BERT scores with PETER.

Under the lexical constraints, the results of existing explanation generation models become lower than the results of aspect-planning generation which indicates current explanation generation models struggle to include specific information (i.e.,~keyphrases) in explanations. Although current lexically constrained generation methods produce text with high diversity, they tend to insert less-related tokens with users and items. Hence, the generated text is less coherent (low n-gram metric results) than \our because these methods cannot incorporate user personas and item profiles from references which are important for explainable recommendation. In contrast, \our easily includes keyphrases in explanations and learns user-item information from references. Therefore, our model largely outperforms existing explanation generation models and lexically constrained generation models. 

Based on the discussion, we argue \our unifies the aspect planning and lexical constraints for explainable recommendations.

\begin{figure}[t]
    \centering
    \includegraphics[width=\linewidth]{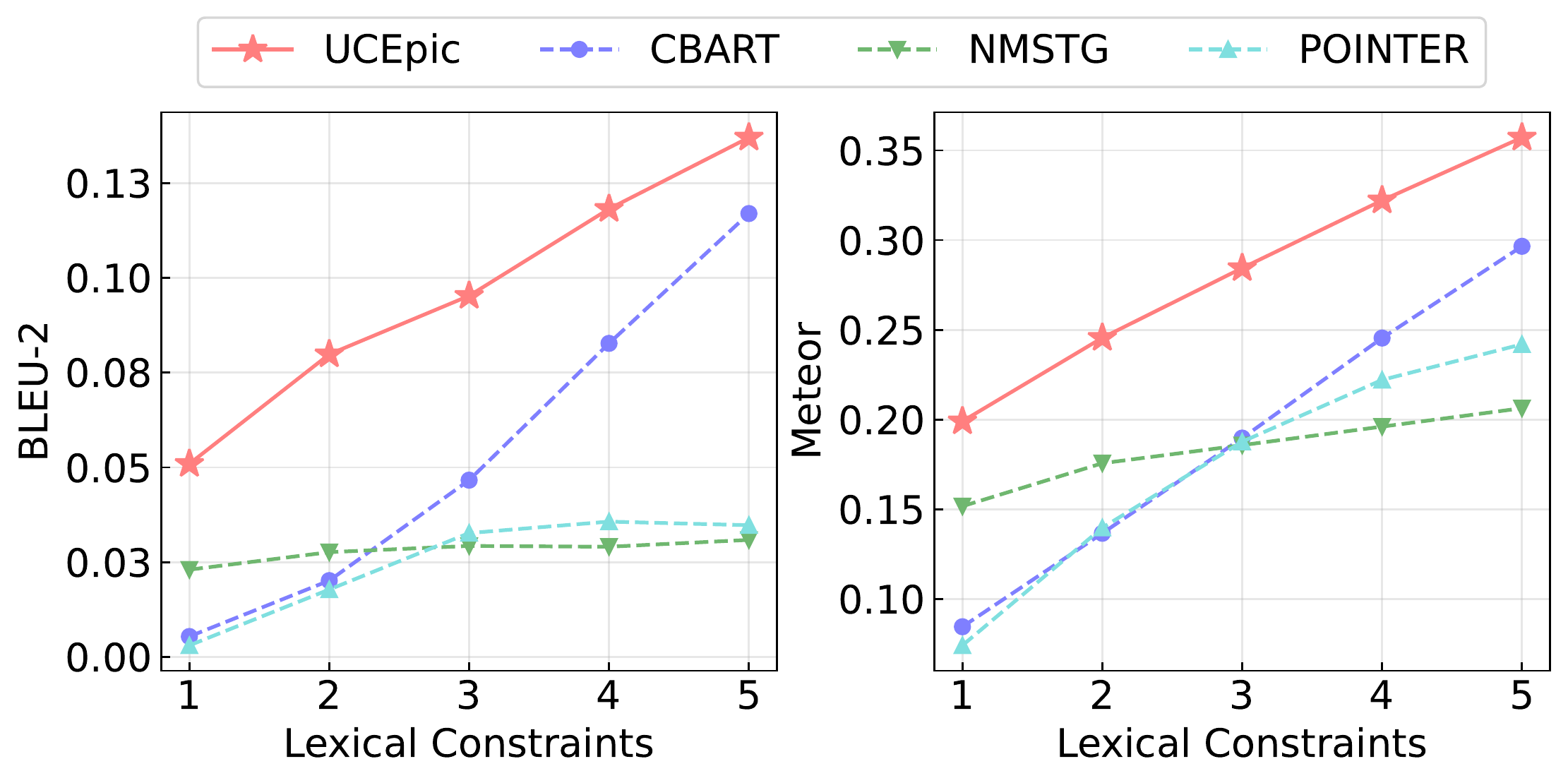}
    \caption{Performance (i.e.,~B-2 and Meteor) of lexically constrained generation models on RateBeer data with different numbers of keyphrases.}
    \vspace{-4mm}
    \label{fig:keywords}
\end{figure}

\subsubsection{Number of Lexical Constraints}
\Cref{fig:keywords} shows the performance of lexically constrained generation models under different keyphrase numbers. Overall, \our consistently outperforms other models under different numbers of lexical constraints. In particular, NMSTG and POINTER do not achieve a large improvement as the number of keyphrases increases because they cannot have random keywords and given phrases are usually broken into words. The gap between \our and CBART becomes large as the number of keyphrases decreases since CBART cannot obtain enough information for explanation generation with only a few keywords, but \our improves this problem by incorporating user persona and item profiles from references. The results indicate existing lexically constrained generation models cannot be applied for explanation generation with lexical constraints.

\subsubsection{Ablation Study}
\begin{figure}[t]
    \centering
    \includegraphics[width=\linewidth]{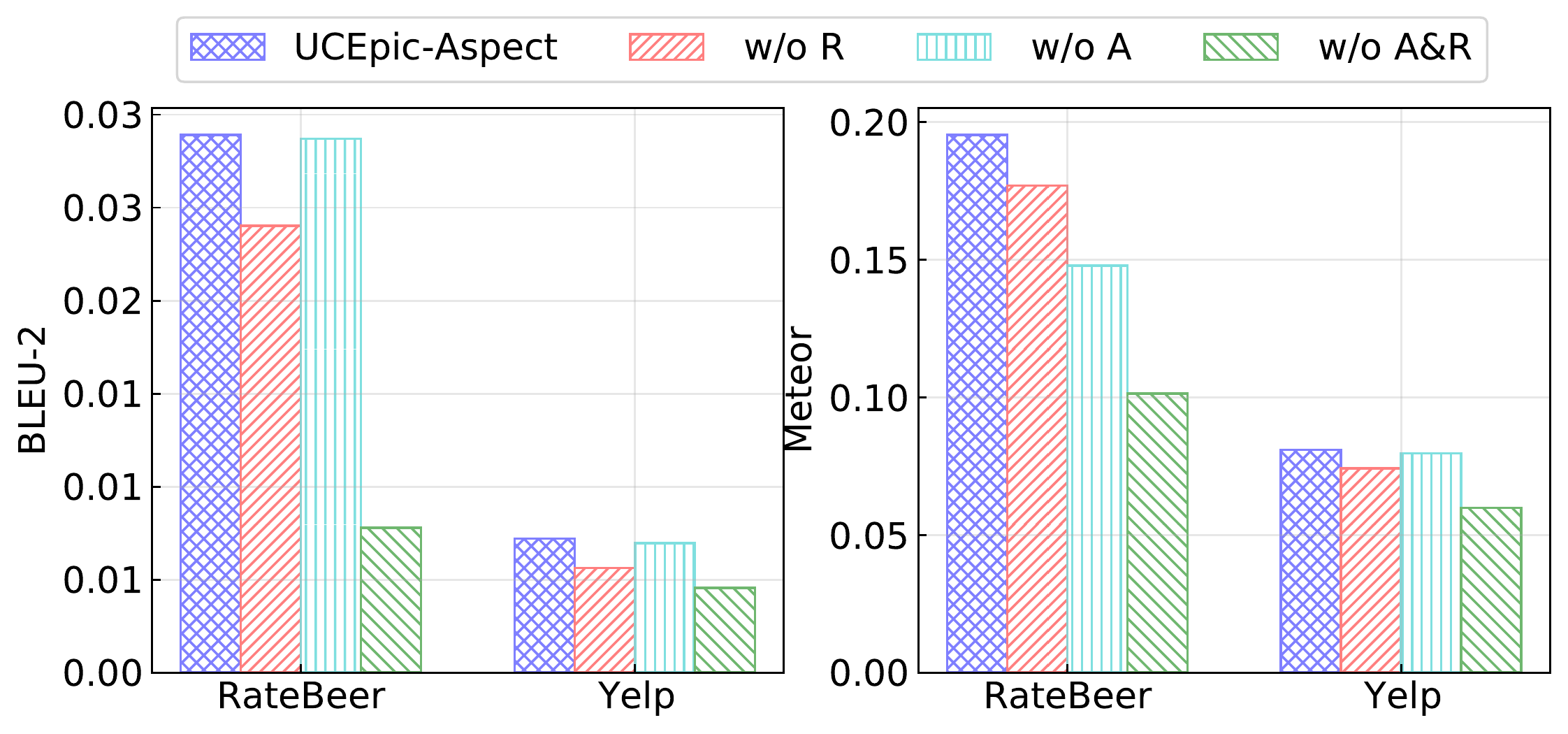}
    \caption{Ablation study on aspects and references.}
    \label{fig:ablation}
    \vspace{-1mm}
\end{figure}

To validate the effectiveness of our unifying method and the necessity of aspects and references for explanation generation, we conduct an ablation study on two datasets and the results are shown in~\Cref{fig:ablation}. We train our model and generate explanations without aspects (w/o A), without references (w/o R) and without both of them (w/o A\&R). From the results, we can see that BLEU-2 and Meteor decrease if we do not give aspects to the model because the aspects can guide the semantics of explanations. Without references, the model generates similar sentences which usually contain high-frequency words from the training data. The performance drops markedly if both references and aspects are absent from the model. Therefore, our unifying method for references and aspects is effective and provides user-item information for explanation generation.

\begin{table}[t]
\centering
\caption{\our with different constraints on Yelp dataset. L denotes lexical constraints.}
\vspace{-1mm}
\scalebox{0.96}{
\begin{tabular}{l|ccccccc}
\toprule
\multicolumn{1}{c|}{\textbf{Constraints}} & \textbf{B-1}   & \textbf{B-2}  & \textbf{D-1}  & \textbf{D-2}   & \textbf{M}     & \textbf{R}     & \textbf{BS}    \\ \midrule
Aspect              & 8.03  & 0.72 & 1.89 & 14.75 & 8.09  & 11.58 & 83.53 \\
L-Extract              & 13.77 & 3.06 & 2.85 & 20.39 & 14.45 & 16.92 & 84.55 \\
L-Frequent             & 10.05 & 0.87 & 2.02 & 15.88 & 9.14  & 12.23 & 83.73 \\
L-Random               & 9.81  & 0.79 & 3.00 & 21.04 & 8.73  & 11.61 & 83.50 \\
Aspect \& L            & 13.12 & 3.01 & 2.89 & 20.34 & 14.41 & 16.94 & 84.56 \\ \bottomrule
\end{tabular}
}
\vspace{-1mm}
\label{tab:constraints}
\end{table}

\begin{table*}[t]
\centering
\caption{Generated explanations from Yelp dataset. Lexical constraints (phrases) are highlighted in explanations.}
\vspace{-2mm}
\scalebox{0.95}{
\begin{tabular}{l|p{7.5cm}p{8.5cm}}
\toprule
\textbf{Phrases}             & \textbf{\hb{pepper chicken}}                                                                                                                                                       &\textbf{ \hb{north shore}, \hb{meat}  }                                                                                                                                                                                                  \\ \cmidrule(l){1-2} \cmidrule(l){3-3}
Human        & Food was great.  The \hb{pepper chicken} is the best.  This place is neat and clean.  The staff  are sweet.  I recomend them to anyone!!                                                                                                   &  Great Italian food on the \hb{north shore}! Menu changes daily based on the ingredients they can get locally. Everything is organic and made "clean". There is no freezer on the property, so you know the \hb{meat} was caught or prepared that day. The chef is also from Italy! I highly recommend!                                                                                                      \\ \cmidrule(l){1-2} \cmidrule(l){3-3}
Ref2Seq             & best restaurant in town ! ! !                                                                                                                   &  what a good place to eat in the middle of the area . the food was good and the service was good .                                                                                                                          \\ \cmidrule(l){1-2} \cmidrule(l){3-3}
PETER               & This place is great! I love the food and the service is always great. I love the \hr{chicken} and the \hr{chicken} fried rice. I love this place.                                      &   The food was good, but the service was terrible. The kitchen was not very busy and the kitchen was not busy. The kitchen was very busy and the kitchen was not busy.                                                                                        \\ \cmidrule(l){1-2} \cmidrule(l){3-3}

POINTER             & \hr{pepper} sauce \hr{chicken} !           &  one of the best restaurants in the \hr{north} as far as i love the south \hr{shore} . great \hb{meat} ! ! \\ \cmidrule(l){1-2} \cmidrule(l){3-3}
CBART             & Great spicy \hr{pepper} buffalo wings and \hr{chicken} wings.           &  Best pizza on the \hb{north shore} ever!  Meatloaf is to die for, especially with \hb{meat} lovers.  \\ \cmidrule(l){1-2} \cmidrule(l){3-3}
\our & Great Chinese restaurant, really great food! The customer service are amazing!  Everything is delicious and delicious! I think this local red hot \hb{pepper chicken} is the best.                                                                                                  &   I had the best Italian \hb{north shore} food. The service is great, \hb{meat} that is fresh and delicious. Highly recommend!                                                           \\ \bottomrule
\end{tabular}
}
\vspace{-2mm}
\label{tab:cases}
\end{table*}

\subsubsection{Kind of Constraints}
We study the performance of \our with different kinds of constraints on the Yelp dataset and the results are shown in~\Cref{tab:constraints}. The settings of Aspect and L-Extract are consistent as \our under aspect-planning and lexical constraints~\footnote{Aspects and phrases are extracted from the generation target and we randomly sampled one aspect and one phrase as model inputs.} respectively in~\Cref{tab:main_perform}. We also study three other kinds of constraints:
\begin{inparaenum}[(1)]
\item L-Frequent. We use the most frequent noun phrase of an item as the lexical constraint.
\item L-Random. We randomly sample the lexical constraint from all noun phrases of an item.
\item Aspect \& L. This method combines both aspect-planning and lexical constraints demonstrated in~\Cref{tab:main_perform} and uses the two kinds of constraints simultaneously.
\end{inparaenum}
From the results, we can see that
\begin{inparaenum}
    \item L-Extract and Aspect \& L have similar results which indicate the lexical constraints have strong restrictions on the generation process hence the aspect planning rarely has controllability on the results.
    \item Generation with lexical constraints can achieve better results than aspect-planning generation.
    \item Lexical constraint selections (i.e.,~L-Extract, L-Frequent, L-Random) result in significant differences in generation performance, which movitate that lexical constraint selections can be further explored in future work.
\end{inparaenum}

\subsection{Human Evaluation}
\begin{figure}[t]
    \centering
    \includegraphics[width=\linewidth]{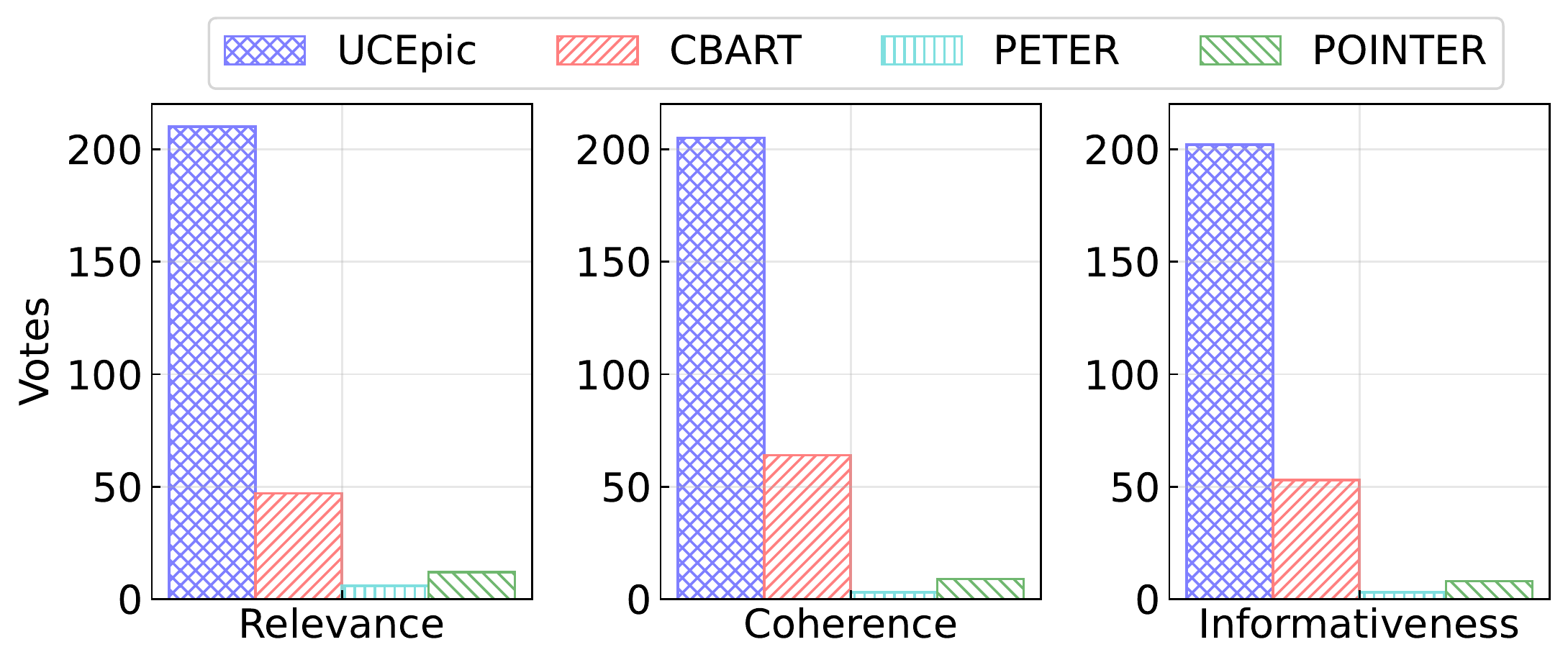}
    \caption{Human evaluation on explanation quality.}
    \label{fig:human}
    \vspace{-4mm}
\end{figure}
We conduct a human evaluation on generated explanations. 
Specifically, We sample 500 ground-truth explanations from Yelp dataset, then collect corresponding generated explanations from PETER-aspect, POINTER, CBART and \our respectively. Given the ground-truth explanation, annotator is requested to select the \emph{best} explanation on different aspects i.e.,~\emph{relevance}, \emph{coherence} and \emph{informativeness}) among explanations generated from PETER. POINTER, CBART and \our (see~\Cref{app:human} for details). We define \emph{relevance}, \emph{coherence} and \emph{informativeness} as:
\begin{itemize}
    \item \textbf{Relevance:} the details in the generated explanation are consistent and relevant to the ground-truth explanations.
    \item \textbf{Coherence:} the sentences in the generated explanation are logical and fluent.
    \item \textbf{Informativeness:} the generated explanation contains specific information, instead of vague descriptions only.
\end{itemize}
The voting results are shown in~\Cref{fig:human}. We can see that \our largely outperforms other methods in all aspects especially for relevance and informativeness. In particular, lexically constrained generation methods (\our and CBART) significantly improve the quality of explanations because specific product information can be included in explanations by lexical constraints. Because POINTER is not robust to random keyphrases, the generated explanations do not get improvements from lexical constraints. 

\subsection{Case Study}
We compare generated explanations from existing explanation generation models (i.e.,~Ref2Seq, PETER), lexically constrained generation models (i.e.,~POINTER, CBART) and \our in~\Cref{tab:cases}. We can see that Ref2Seq and PETER usually generate general sentences which are not informative because they struggle to contain specific item information by traditional auto-regressive generation. POINTER and CBART can include the given phrases (pepper chicken) in their generation, but they are not able to learn information from references and hence generate some inaccurate words (pepper sauce chicken, chicken wings) which mislead users. In contrast, \our can generate coherent and informative explanations which include the specific item attributes and are highly relevant to the recommended item.

%% file: 5_conclusion.tex
\section{Conclusion}
In this paper, we propose to have lexical constraints in explanation generation which can largely improve the informativeness and diversity of generated reviews by including specific information. To this end, we present \our, an explanation generation model that unifies both aspect planning and lexical constraints in an insertion-based generation framework. We conduct comprehensive experiments on RateBeer and Yelp datasets. Results show that \our significantly outperforms previous explanation generation models and lexically constrained generation models. Human evaluation and a case study indicate \our generates coherent and informative explanations that are highly relevant to the item. 

%% file: 6_appendix.tex
\section{Motivating Experiment Details}
\label{app:motivation}
In this experiment, we evaluate the diversity and informativeness of explanations. Specifically, we apply phrase coverage, aspect coverage and Distinct-2 to measure generated explanations and human-written explanations.

For \textbf{phrase coverage}, we extract noun phrases from explanations by spaCy noun chunks. Then we compare the phrases in human-written explanations and generated explanations. If a phrase appears in both explanations, we consider it as a covered phrase by generated explanations. This experiment measures how much specific information can be included in the generated explanations.

For \textbf{aspect coverage}, we use the aspect extraction tool~\cite{Li2022UCTopicUC} per dataset to construct a table that maps phrases to aspects, then we map the phrases in generated explanations to aspects by looking up the phrase-aspect table. For each sample, we calculate how many aspects in ground-truth explanation are covered in generated explanations and report the average aspect coverage per dataset.


For \textbf{Distinct-2}, we use the numbers as described in~\Cref{tab:main_perform}.

\section{Baseline Details}
\label{app:baseline}

For \textbf{ExpansionNet}, we use the default setting which uses hidden size 512 for RNN encoder and decoder, batch size as 25 and learning rate 2e-4. For aspect planning in ExpansionNet, we use the set of lexical constraints (as concatenated phrases) to replace the \emph{title} or \emph{summary} input as contextual information for training and testing. 

For \textbf{Ref2Seq}, we use the default setting with 256 hidden size, 512 batch size and 2e-4 learning rate. For aspect planning, we concatenate our given phrases as references (historical explanations are also incorporated as references following the original implementation) as contextual information in training and testing.

For \textbf{PETER}, we use the original setting with 512 embedding size, 2048 hidden units, 2 self-attention heads with 2 transformer layers, 0.2 dropout. We use the training strategy suggested by the authors. Since original PETER only supports single words as an aspect, we adopt PETER to multiple words with a maximum length of 20 and reproduce the original single-word model on our multi-word model. We input our lexical constraints as the multi-word input for PETER training and testing.

For \textbf{NMSTG}, we use the default settings with an LSTM with 1024 hidden size with the uniform oracle. We convert our lexical constraints into a prefix sub-tree as the input of NMSTG, and then use the best sampling strategy in our testing (i.e., \texttt{StochasticSampler}) for NMSTG.

For \textbf{POINTER}, we use the pre-training \texttt{BERT-large}~\cite{Devlin2019BERTPO} (\#params $\approx$ 340M.) from WIKI to fine-tune 40 epochs on our downstream datasets. We use all the default settings except batch sizes since POINTER requires 16 GPUs for distributed training that exceeds our computational resources. Instead, we train POINTER with the same configuration on 3 GPUs. For testing, we select the base maximum turn as 3 with the default greedy decoding strategy. We feed lexical constraints as the original implementation.

For \textbf{CBART}, we use the checkpoint pre-trained on \texttt{BERT-large}~\cite{Devlin2019BERTPO} (\#params $\approx$ 340M.) with the one-billion-words dataset to fine-tune our downstream datasets. We use the `tf-idf' training mode and finetune it on one GPU. For testing, we select the greedy decoding strategy. We set other hyper-parameters to default as the code base~\footnote{https://github.com/NLPCode/CBART}.

\section{Human Evaluation Details}
\label{app:human}

We conduct human evaluation experiments on Yelp datasets to evaluate the generation quality of generated explanations in terms of \textit{relevance}, \textit{coherence} and \textit{informativeness}. 

\begin{figure}[t]
    \centering
    \includegraphics[width=\linewidth]{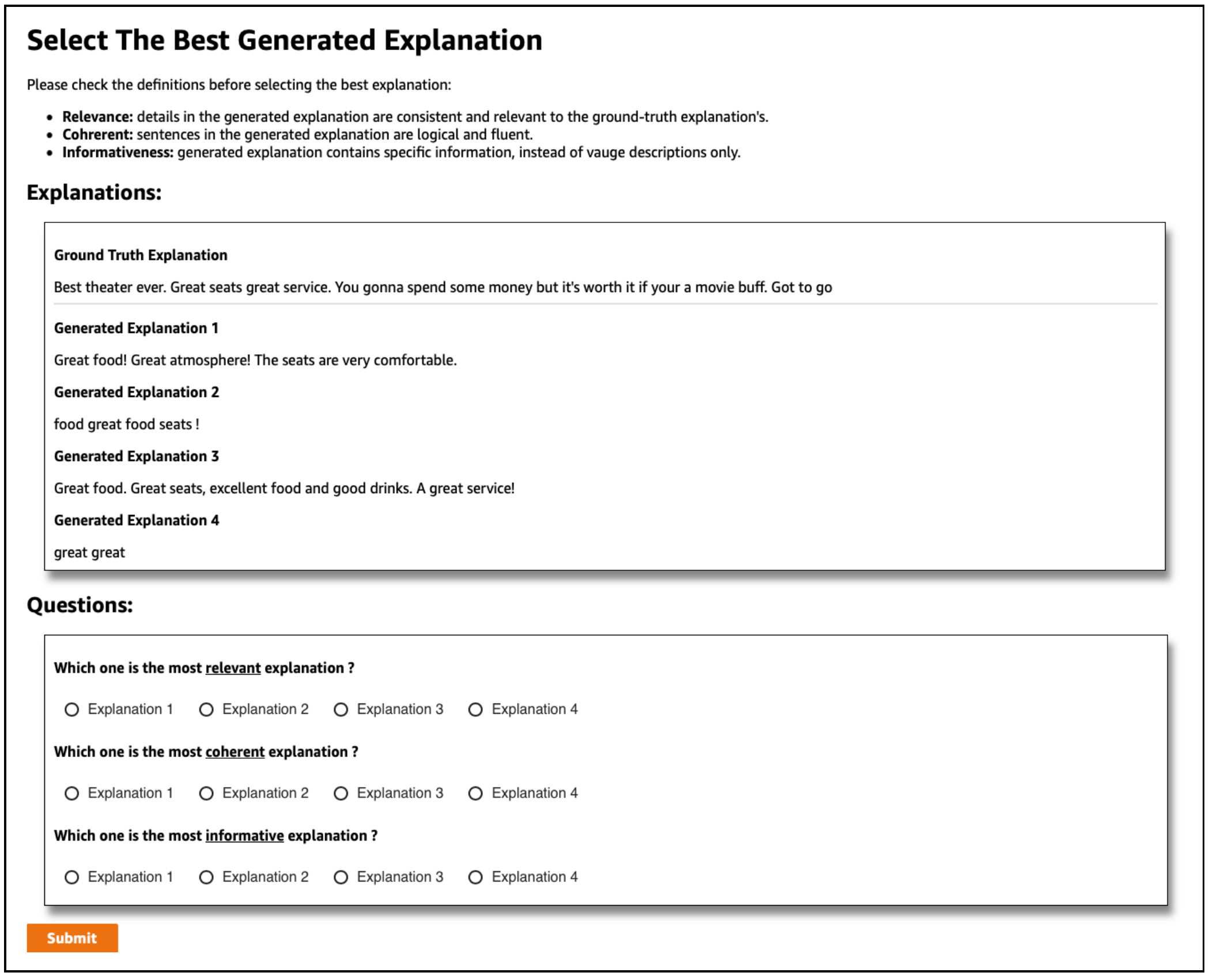}
    \caption{Our human evaluation example on MTurk.}
    \label{fig:example}
\end{figure}

We submit our task to MTurk~\footnote{\url{https://www.mturk.com}} and set the reward as \$0.02 per question. For each question, we first show the definition of \emph{relevance}, \emph{coherence} and \emph{informativeness}, then we shuffle the order of model-generated explanations to eliminate the positional bias. Each question is requested to be answered by 3 different MTurk workers, who are required to have great than 80\% \texttt{HIT Approval Rate} to improve the quality of answers. \Cref{fig:example} is an example of our evaluation template. We collect the answers and count the majority votes, where the majority vote is defined as model $i$ has 2 or more votes (since we have 3 answers per question). We ignore the questions without majority votes. Finally, we collected 1,120 valid votes for 370 questions, in which 275 \textit{relevance} questions, 281 \textit{coherence} questions and 266 \textit{informativeness} questions have majority votes.
